\documentclass[conference]{IEEEtran}
\IEEEoverridecommandlockouts

\usepackage{cite}
\usepackage{amsmath,amssymb,amsfonts}
\usepackage{booktabs}
\usepackage{algorithm}
\usepackage{algpseudocode}
\usepackage{graphicx}
\usepackage{textcomp}
\usepackage{xcolor}
\usepackage{multirow}
\usepackage{tipa}
\usepackage{placeins}
\usepackage{hyperref}

\def\BibTeX{{\rm B\kern-.05em{\sc i\kern-.025em b}\kern-.08em
    T\kern-.1667em\lower.7ex\hbox{E}\kern-.125emX}}
\begin{document}

\title{Data-Driven Mispronunciation Pattern Discovery for Robust Speech Recognition}

\author{\IEEEauthorblockN{Anna Seo Gyeong Choi $^*$}
\IEEEauthorblockA{\textit{Cornell University}\\
Ithaca, New York, USA \\
sc2359@cornell.edu}
\and
\IEEEauthorblockN{Jonghyeon Park $^*$}
\IEEEauthorblockA{\textit{NAVER Cloud Corporation} \\
Republic of Korea \\
jong-hyeon.park@navercorp.com}
\and
\IEEEauthorblockN{Myungwoo Oh}
\IEEEauthorblockA{\textit{NAVER Cloud Corporation} \\
Republic of Korea \\
myungwoo.oh@navercorp.com}
}

\maketitle
\def\thefootnote{*}\footnotetext{These authors contributed equally to this work}\def\thefootnote{\arabic{footnote}}

\begin{abstract}

Recent advancements in machine learning have significantly improved speech recognition, but recognizing speech from non-fluent or accented speakers remains a challenge. Previous efforts, relying on rule-based pronunciation patterns, have struggled to fully capture non-native errors. We propose two data-driven approaches using speech corpora to automatically detect mispronunciation patterns. By aligning non-native phones with their native counterparts using attention maps, we achieved a 5.7\% improvement in speech recognition on native English datasets and a 12.8\% improvement for non-native English speakers, particularly Korean speakers. Our method offers practical advancements for robust Automatic Speech Recognition (ASR) systems particularly for situations where prior linguistic knowledge is not applicable.

\end{abstract}


\section{Introduction}

Advances in ASR have yielded impressive results, particularly in controlled environments with abundant training data. However, recognizing speech from non-fluent speakers or those with non-native accents remains a significant challenge due to the limited availability of diverse datasets and under-explored algorithmic approaches \cite{b1}. Accurately transcribing speech from non-native speakers often requires scaling up both data and model size, which is infeasible with limited resources, particularly when there is a lack of information on the speakers’ native language (L1). Simply augmenting data without accounting for underlying pronunciation variations has proven insufficient in improving ASR performance for non-native (L2) speech \cite{b2}. In addition to transcription, analyzing pronunciation features is crucial to developing robust speech AI systems, particularly in fields like language education. However, end-to-end ASR models often fall short in performing detailed pronunciation-level analyses, limiting their applicability in domains that require such insights. 

To address these challenges, we propose a method for automatically identifying mispronunciations in non-native speech to enhance pronunciation learning. By leveraging a model trained on native speaker data, we extract pronunciation sequences from non-native English speech spoken by Koreans and use attention mechanisms to identify word-level pronunciation error patterns. This data-driven approach allows us to improve ASR performance by 5.7\% on native English datasets and a 12.8\% for non-native English speakers, particularly Korean speakers.

\section{Previous Research}

Incorporating pronunciation modeling for lexicon expansion has been the focus of several studies, particularly through acoustic data-driven approaches aimed at enhancing ASR systems.Lu, Ghoshal, and Renals \cite{b3} investigated both Expectation Maximization and Viterbi-based algorithms to automatically develop pronunciation lexicons, demonstrating that these hand-crafted approaches significantly improve accuracy compared to grapheme-to-phoneme (G2P) transformations. Zhang et al. \cite{b4} extended these ideas by introducing a greedy framework for pronunciation selection based on acoustic evidence and a likelihood-reduction approach, addressing the issue of pruning lexicons to balance the trade-off between pronunciation variant coverage and ASR performance. Chen, Povey, and Khudanpur \cite{b5} explored similar acoustic data-driven pronunciation modeling for logographic languages where G2P models struggle with unseen characters, using iterative framework to train G2P models capable of generating pronunciations for previously unseen graphemes.

Our previous work \cite{b6} explored a rule-based approach to identifying mispronunciation patterns by leveraging integrated phonetic similarities and differences between Korean and English, which involved manually crafting rules based on linguistic knowledge to map common mispronunciations which frequently occur in Korean speakers of English. These rules were then incorporated into a multi-lexicon ASR framework, which allowed us to explicitly model expected non-native pronunciation errors. This method systematically applied predefined phonological rules to create a comprehensive lexicon covering all possible variations of mispronunciations for each phoneme. The rule-based lexicon was designed to account for the full spectrum of potential non-native pronunciation errors when the L1 is fixed and there is ample information known about the relationship between the two languages, L1 and L2. While this method proved effective in the research, real-life applications of ASR systems call for situations where the L1 is unknown, the crosslinguistic knowledge is understudied, or there are too many L1s to create a rule-based lexicon. To mitigate this limitation, we incorporated a direct data-driven methodology where instead of relying solely on a set of pre-defined rules, we automatically extract pronunciation variants directly from the non-native speech corpus. This approach leads to a more concise lexicon, containing only the pronunciation variants that actually appear in the data, allowing for a more targeted and adaptive ASR model, as well as solving the cases where L1 information is missing.

Large-scale pre-trained models like Whisper \cite{b18} have achieved breakthroughs in general ASR performance. However, their performance often deteriorates significantly for in-domain non-native speech when pronunciation diversity is not adequately addressed. Compared to domain-adapted models like LibriSpeech-fitted ones, such models struggle to maintain robustness, highlighting the importance of targeted pronunciation modeling in enhancing ASR systems.

\section{Proposed Methods}

\begin{figure}[t]
\centering
\includegraphics[width=\linewidth]{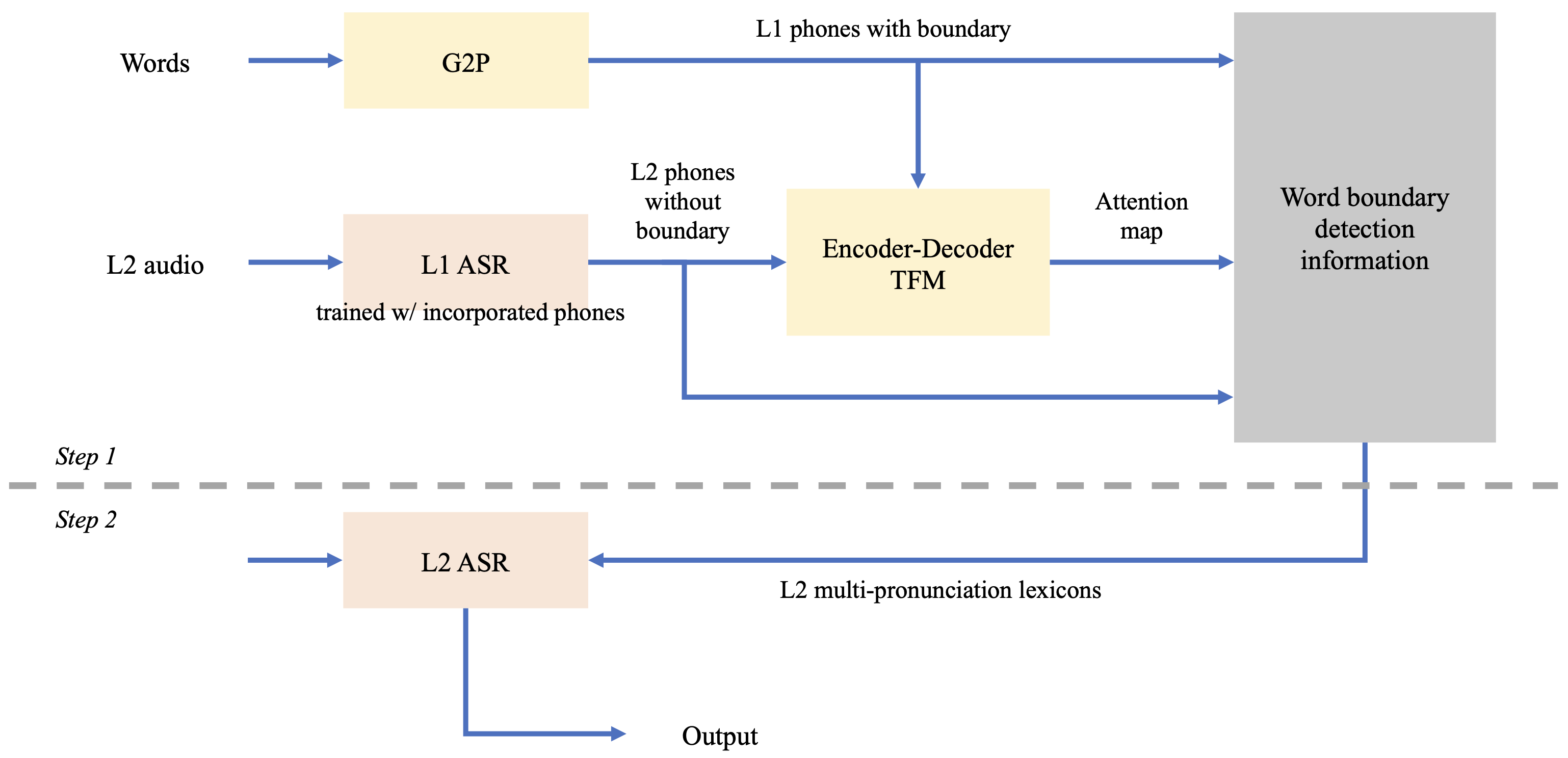}
\caption{Pipeline for extracting mispronunciation pattern. Non-native speech input is processed through an ASR model trained on native data, followed by pseudo phone labeling and alignment. Lexicon information is then incorporated for ASR performance enhancement.}
\label{fig:blob}
\end{figure}

The overall pipeline of our method is represented in Fig \ref{fig:blob}. We input English data spoken by native Korean speakers through an ASR model trained only on L1 speech, utilizing both English and Korean phones. Using the L1 ASR model that is made of a unigram phoneme language model, we extract pseudo phone labels without any word boundaries as output,during which process we specifically apply boosting on Korean native phones by 7.9 times in the unigram G model to increase the chances of including non-native mispronunciation patterns. We then align these pseudo phones to the reference phones created using an English G2P converter. The lexicon information collected from the alignments are then re-incorporated to the ASR model for performance enhancement.
Since our non-native phone sequences do not have word boundaries, the purpose of the alignment is to form the phones into corresponding native words, during which process the multi-pronunciation lexicons are created due to disagreement in placing the word boundaries. We compare two novel methods for this: (1) dynamic programming-based and (2) attention-based. 
Needleman-Wunsch \cite{b7} is a dynamic programming algorithm that finds an optimal alignment between two sequences by minimizing the cost of mismatches, insertions, and deletions. The algorithm has been used for speech alignment in past literature \cite{b8} \cite{b9}, but our research is the first to use it for the purpose of lexicon creation. The algorithm creates a scoring matrix with dimensions $(n+1) \times (m+1)$, where $n$ and $m$ are the lengths of the non-native and native sequences, respectively. The matrix is filled by calculating the score at each cell based on three possible operations: match/mismatch, insertion and deletion. Once the matrix is filled, the optimal alignment is determined by backtracking from the bottom-right corner of the matrix to the top-left. During this process, the algorithm reconstructs the aligned sequences, including any gaps introduced during the alignment. We create a lexicon based on the word boundary placed by this alignment. See Algorithm \ref{alg:needleman_wunsch} below \footnote{Code to reproduce the algorithm are available on \href{https://github.com/AnnaSeoGyeongChoi/data-driven-pronunciation-2024}{Github}.}.

\begin{algorithm}
\caption{Dynamic programming-based lexicon expansion}
\label{alg:needleman_wunsch}
\begin{algorithmic}[1]
\Function{NeedlemanWunsch}{seq1, seq2, match\_score, mismatch\_score, gap\_penalty}
    \State Initialize score matrix of size (len(seq1) + 1) $\times$ (len(seq2) + 1)
    
    \For{each i from 1 to len(seq1)}
        \State score[i][0] $\gets$ score[i-1][0] + gap\_penalty
    \EndFor
    \For{each j from 1 to len(seq2)}
        \State score[0][j] $\gets$ score[0][j-1] + gap\_penalty
    \EndFor
    
    \For{each i from 1 to len(seq1)}
        \For{each j from 1 to len(seq2)}
            \State match $\gets$ score[i-1][j-1] + (match\_score if seq1[i-1] = seq2[j-1] else mismatch\_score)
            \State delete $\gets$ score[i-1][j] + gap\_penalty
            \State insert $\gets$ score[i][j-1] + gap\_penalty
            \State score[i][j] $\gets$ min(match, delete, insert)
        \EndFor
    \EndFor
    
\EndFunction
\end{algorithmic}
\end{algorithm}

\begin{figure}[t]
\centering
\includegraphics[width=\linewidth]{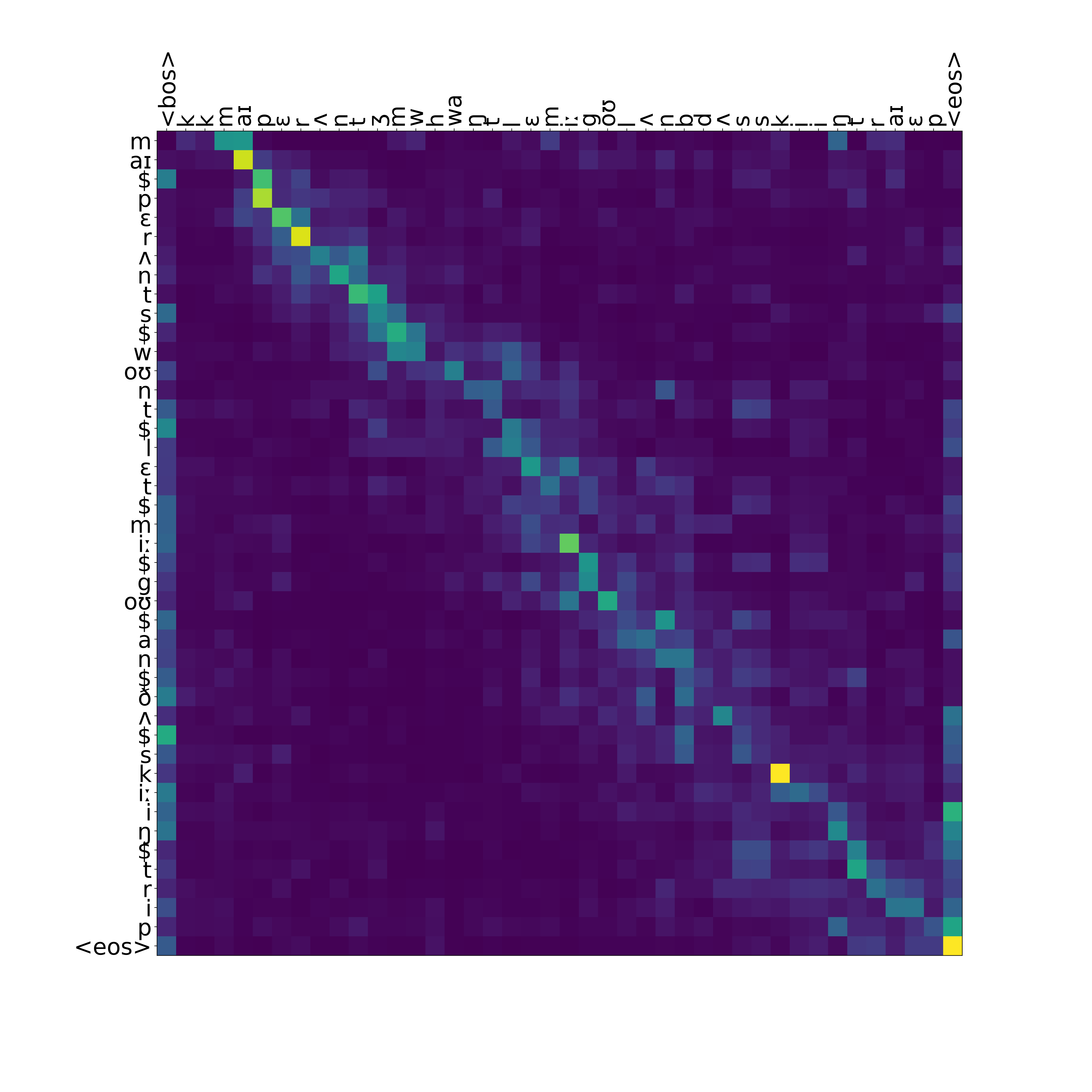}
\caption{Example of an attention map information used in the attention-based approach. On the top of the map are non-native phones without word boundaries, and on the left side are corresponding native phones with word boundaries.}
\label{fig:attention}
\end{figure}

Our attention-based lexicon leverages the attention mechanisms in the encoder-decoder transformer model \cite{b10} -- we extract maximum attention values from the final layer of the model, which help us identify the most likely word boundaries by pinpointing the positions in the native phone sequences. Once the maximum attention positions are identified from the attention maps, as exemplified in Fig \ref{fig:attention}, we generate pronunciation variants by shifting the word boundaries around these positions. We consider three possible phone distances from the original maximum attention point to account for variations in pronunciation timing and attention framing. The generated variations are then compared with the reference native sequences using edit distance calculations to evaluate the phonemic similarity and select the most accurate alignment. The algorithm is shown on Algorithm \ref{alg:attention}.

Attention information is yet to be incorporated in the lexicon creation pipeline. There is very limited research in using attention mechanism for lexicon expansion \cite{b11} and our work is the first in that we not only use the attention information, but we also employ a sort of beam search method to make up for the possibility that the attention information is inherently faulty.

\begin{algorithm}
\caption{Attention-based lexicon expansion}
\label{alg:attention}
\begin{algorithmic}[1]
\Function{AlignWordBoundaries}{input\_data, n}
    \For{each (non\_native, native) \textbf{in} input\_data}
        \State variations $\gets$ SplitByAttentionMap(non\_native, n)
        \State distances $\gets$ []
        
        \For{var \textbf{in} variations}
            \State dist $\gets$ ComputeEditDistance(var, native)
            \State Append(distances, dist)
        \EndFor
        \State best\_var $\gets$ SelectMin(distances)
        \If{best\_var $\leq$ threshold} 
            \State Return best\_var
        \EndIf
    \EndFor
\EndFunction

\Function{SplitByAttentionMap}{non\_native, n}
    \State phones $\gets$ Split\_by\_attention\_map(non\_native)
    \State variations $\gets$ []
    
    \For{shift \textbf{from} -n \textbf{to} n}
        \State shifted\_phones $\gets$ ApplyBoundaryShift(phones, shift)
        \State Append(variations, shifted\_phones)
    \EndFor
    
    \State \Return variations
\EndFunction
\end{algorithmic}
\end{algorithm}

\begin{table*}[ht!]
\caption{WER (\%) of evaluation sets.}
\label{tab:wer_ft}
\centering
\begin{tabular}{c|c|ccccc|ccc}
\toprule
\multirow{2}{*}{\textbf{\#}} & \multirow{2}{*}{\textbf{Lexicon type}} & \multicolumn{5}{c|}{\textbf{L1 testsets}} & \multicolumn{2}{c}{\textbf{L2-Arctic testsets}} \\
& & \textbf{dev-clean} & \textbf{dev-others} & \textbf{test-clean} & \textbf{test-others} & \textbf{CMU-ARCTIC} & \textbf{KR} & {\textbf{Others}} \\
\midrule
1 & no mispronun & 3.94 & 8.95 & 4.44 & 9.47 & 2.81 & 9.56 & 17.16 \\
2 & rule-based & \textbf{3.74} & \textbf{7.84} & \textbf{4.11} & \textbf{8.36} & \textbf{2.48} & 8.59 & 15.60\\
\midrule
3 & dp align & 4.02 & 8.83 & 4.36 & 9.33 & 2.84 & 10.35 & 17.71 \\
4 & attention align & 3.81 & 8.11 & 4.17 & 8.44 & 2.58 & 8.71 & 15.89 \\
5 & rule + attention align & 3.77 & 7.99 & 4.13 & 8.44 & 2.53 & \textbf{8.34} & \textbf{14.90} \\
\bottomrule
\end{tabular}
\end{table*}

\section{Experiments}
\subsection{Datasets}
For this study, we initialized our model using the one developed in our prior research \cite{b6}. The model was trained on a unified phoneme set consisting of 62 phonemes, incorporating Korean L1 articulatory features, to handle both native and non-native pronunciations more effectively. The training data consisted of 960 hours of English speech from the LibriSpeech (LS) dataset \cite{b12} and 960 hours of Korean speech from the KSponSpeech dataset \cite{b13}, yielding a total of 1,920 hours of L1 data. To extend the model's capacity to handle non-native speech, we incorporated an additional 1,000 hours of in-house L2 English speech data EngDictKr, collected via crowd-sourcing from Korean speakers with varying levels of English proficiency. This dataset, being the ony L2 speech data used for training, was also used to extract mispronunciation information for the lexicon incorporation. We also trained a baseline model using the conventional English phoneme set of 39 phonemes, without the integration of Korean phone features developed in our prior research. This allowed us to directly compare the impact of solely data-driven lexicon expansion on the recognition performance.

For evaluation purposes, we employed multiple test sets. For L1 English speech, we used the LibriSpeech dataset, including both the dev-clean/other and test-clean/other subsets. The LibriSpeech test sets contain 5.39/5.12 hours of speech in the dev-clean/other subset and 5.34/5.40 hours in the test-clean/other subset, with 2,703/2,864 and 2,620/2,939 utterances, respectively. Additionally, we used the CMU-ARCTIC corpus \cite{b14} for both L1 and L2 evaluations, which consists of 4 hours of speech and 4,524 utterances. For L2 English speech, we incorporated two test sets: the L2-ARCTIC-Kr dataset \cite{b15}, containing 4 hours of speech and 4,524 utterances, and the L2-ARCTIC-other dataset \cite{b15}, which includes 22,323 utterances spanning 22.5 hours of speech from non-Korean L1 speakers. These datasets allowed us to thoroughly evaluate the model's performance across both native and non-native speech, with a specific focus on improving recognition accuracy for L2 English spoken by Korean speakers.

\subsection{Experimental setup}
We employed the same ASR model architecture as in our prior research \cite{b6}, specifically utilizing a conformer encoder architecture \cite{b16} without a decoder network. The model consisted of 26.8 million parameters, comprising 16 layers with 256 dimensions and 8 multi-head attention mechanisms. The input features were 80-dimensional filterbank features, which were computed from 25 ms windows with a 10 ms step size. These features were then passed through convolutional subsampling layers, reducing the time resolution to a 40 ms output rate. For encoder-decoder transformer modeling, we employed a dual-encoder and four-decoder setup. The embedding size was set to 512, with 8 attention heads and a feed-forward network (FFN) dimension of 512. The final output size was 66, which included 62 phoneme units as well as additional special tokens required for processing. The models were trained using the lattice-free maximum mutual information (LF-MMI) objective \cite{b17}, which allows for the simultaneous modeling of multiple pronunciation candidates. We evaluated the performance of the models on both L1 and L2 test sets, enabling us to assess the impact of the proposed phoneme integration on native and non-native speech recognition.

\section{Results}

The results of our experiments are summarized in Table \ref{tab:wer_ft}. The baseline model (\#1) that does not incorporate any mispronunication handling exhibits the highest WER -- worst performance -- for all of L1 and L2 test sets, and all the models with mispronunciation handling, regardless of the alignment information type, show significant improvements from model \#1, especially so for non-native speech.

For the L1 test sets, the combined rule-based and attention-based model (\#5) indeed performs the best overall, exceeding expectations and indicating complementary benefits from combining the two approaches. While the dynamic programming-based model (\#3) perform subpar than the rule-based baseline model (\#2), the attention-based model's (\#4) performance is almost approaching that of model \#2's performance. Notably, \#4 achieves this performance using far fewer mispronunciation patterns -- total lexicon entry of 35,204 compared to 336,882 for model \#2 and 345,489 for model \#5, which is nearly a 90\% drop in lexicon size. This suggests a more efficient search space compared to the rule-based approach, which is particularly valuable as an increase in lexicon size often results in a larger search space. For the L2-Arctic test sets, substantial gains were observed when using model \#3. The baseline models \#1 and \#2 showed a WER of 9.56\% and 8.59\% for the Korean subset, and 17.16\% and 15.60\% for the non-Korean L1 subsets. While models \#3 and \#4 also provide comparable rates of WER for these subsets, our comprehensive model \#5 further improved L2 speech recognition and reduced the WER to 8.34\% for Korean L1 speakers and 14.90\% for non-Korean L1 speakers. Qualitative analysis of the lexicon reveals linguistically meaningful mispronunciation patterns, particularly for Korean learners of English. For example, the entry \textit{doesn't}, which in the native phoneme sequence is \textipa{[d2znt]}, also includes the alternative \textipa{[d2snt]}, reflecting devoicing of \textipa{[z]}. Similarly, the word \textit{everything} is assigned alternatives such as \textipa{[EvrisIN]} and \textipa{[EbriTIN]}, capturing the common Korean tendencies to tensify \textipa{[T]} and devoice \textipa{[v]}.

Overall, the proposed methods offer significant improvements in ASR performance for both native and non-native speech. The combination of rule-based and attention-based alignment proved particularly effective for L2 speech, where mispronunciations are more frequent. Even though our model was trained only on mispronunciation data from Korean L1 speakers, it achieved improvements for non-Korean L1 test sets as well, suggesting the robustness of the data-driven approach when L1 information is unknown or diverse. Finally, while the vocabulary size extracted from EngDictKr was limited, the performance improvements of models \#2 and \#4 demonstrate that lexicon expansion methods can achieve results comparable to those obtained with larger vocabularies. This highlights the flexibility and efficiency of our approach, especially in cases where L1 mispronunciation patterns are unknown.

\section{Conclusion}

In this study, we proposed a data-driven approach to identifying mispronunciation patterns for non-native speech, specifically for Korean L1 speakers of English, to improve ASR robustness. Our method integrates both dynamic programming and attention-based mechanisms for lexicon creation, and the results indicate substantial improvements in WER specifically for L2 test sets. By automatically discovering mispronunciation patterns from non-native speech, our approach complements traditional rule-based methods, enhancing their performance without requiring extensive linguistic knowledge in the cases where L1 is unknown or multiple and achieves comparable performances with state-of-the-art models like Whisper \cite{b18}, despite being trained on significantly smaller datasets.

Our findings show the importance of incorporating both linguistic knowledge and data-driven discovery for robust speech recognition systems, especially in non-native speech scenarios. Future work could extend this approach to other languages and L1 backgrounds, enhancing the ability of ASR systems to handle diverse accents and pronunciation patterns efficiently. Additionally, the integration of self-supervised learning techniques and more extensive datasets may further boost performance for both native and non-native speech recognition tasks.

\end{document}